\documentclass{article}

\usepackage{arxiv}

\usepackage[utf8]{inputenc} 
\usepackage[T1]{fontenc}    
\usepackage{hyperref}       
\usepackage{url}            
\usepackage{booktabs}       
\usepackage{amsfonts}       
\usepackage{nicefrac}       
\usepackage{microtype}      
\usepackage{lipsum}		
\usepackage{graphicx}
\usepackage{natbib}
\usepackage{doi}

\usepackage{fancyvrb}
\usepackage{xcolor}
\usepackage{booktabs}
\usepackage{tablefootnote}
\usepackage{multirow}
\usepackage{refcount}

\usepackage{todonotes}

\newcommand{\sysname}{GenRL}
\usepackage{amssymb}
\usepackage{pifont}
\newcommand{\cmark}{\ding{51}}%
\newcommand{\xmark}{\ding{55}}%

\newcommand{\repeatthanks}{\textsuperscript{\thefootnote}}

\title{Generative Relation Linking for Question Answering over Knowledge Bases}

\author{ Gaetano Rossiello\thanks{Equal contributions} \And
Nandana	Mihindukulasooriya\repeatthanks \And
Ibrahim	Abdelaziz \And
Mihaela Bornea \And
Alfio Gliozzo \And
Tahira Naseem \And
Pavan Kapanipathi \And \\ IBM Research, T.J. Watson Research Center, Yorktown Heights, NY, USA
}

\date{}

\renewcommand{\shorttitle}{Generative Relation Linking for Question Answering over Knowledge Bases}

\hypersetup{
pdftitle={Generative Relation Linking for Question Answering over Knowledge Bases},
pdfauthor={Rossiello et al.},
pdfkeywords={Relation Linking, Question Answering, Knowledge Bases},
}

\begin{document}
\maketitle

\begin{abstract}
Relation linking is essential to enable question answering over knowledge bases. Although there are various efforts to improve relation linking performance, the current state-of-the-art methods do not achieve optimal results, therefore, negatively impacting the overall end-to-end question answering performance. In this work, we propose a novel approach for relation linking framing it as a generative problem facilitating the use of pre-trained sequence-to-sequence models.
We extend such sequence-to-sequence models with the idea of infusing structured data from the target knowledge base, primarily to enable these models to handle the nuances of the knowledge base. Moreover, we train the model with the aim to generate a structured output consisting of a list of argument-relation pairs, enabling a knowledge validation step. We compared our method against the existing relation linking systems on four different datasets derived from DBpedia and Wikidata. Our method reports large improvements over the state-of-the-art while using a much simpler model that can be easily adapted to different knowledge bases.  
\end{abstract}


\section{Introduction}
\label{sec:intro}

\begin{figure}[t]
  \centering
    \includegraphics[width=0.95\textwidth]{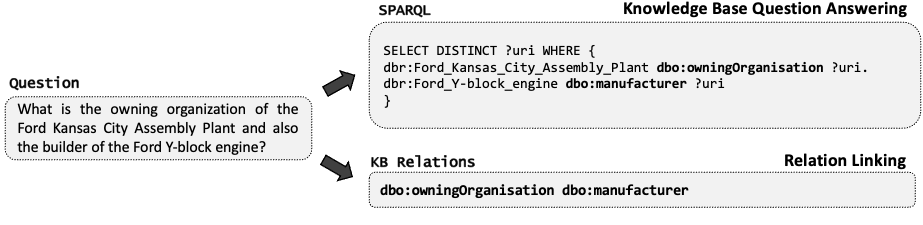}
    \caption{An example taken from LC-QuAD 1.0 showing the difference between KBQA and RL tasks. Knowledge Base Question Answering (on the top): given the question, predict the gold SPARQL query. Relation Linking (on the bottom): given the question, predict the KB relations \textit{dbo:owningOrganisation}, \textit{dbo:manufacturer}.}
    \label{fig:kbqa}
\end{figure}

The goal of Knowledge Base Question Answering (KBQA) systems is to transform natural language questions into SPARQL queries that are then used to retrieve answer(s) from the target Knowledge Base (KB). Relation linking is a crucial component in building KBQA systems. It identifies the relations expressed in the question and maps them to the corresponding KB relations. For example, in Figure~\ref{fig:kbqa}, to translate the question \textit{``What is the owning organization of the Ford Kansas City Assembly Plant and also the builder of the Ford Y-block engine?''} into its corresponding SPARQL query, it is necessary to determine the two KB relations: \textit{dbo:owningOrganisation}, \textit{dbo:manufacturer}.

Relation linking has proven to be a challenging problem, with state-of-the-art approaches performing less than 50\% F1 on the majority of the datasets~\cite{sakor2019old,lin2020kbpearl,mihindukulasooriya2020leveraging}, thus making it a bottleneck for the overall performance of KBQA systems. The challenges primarily arise from the following factors: 
1) relations in text and the KB are often lexicalized differently (implicit mentions); 2) questions with multiple relations and 3) training data is often limited. While past approaches have tried to tackle these issues by either creating hand-coded rules~\cite{sakor2020falcon}, or by using semantic parsing~\cite{mihindukulasooriya2020leveraging}, these challenges can be naturally addressed using the latest advances in auto-regressive sequence-to-sequence models (seq2seq) which have been shown to perform surprisingly well on tasks such as question answering~\cite{DBLP:conf/nips/LewisPPPKGKLYR020}, slot filling~\cite{DBLP:journals/corr/abs-2009-02252} or entity linking~\cite{DBLP:journals/corr/abs-2010-00904}, in a generative fashion. However, seq2seq models have not yet been explored for relation linking, particularly in the context of KBQA. In this work, we introduce \sysname, a novel generative approach for relation linking that capitalises on pre-trained seq2seq  models.

A simple seq2seq model for relation linking can be trained using just the question text to generate a sequence of relations. However, such models,  trained on only the question text, are unable to deal with the nuances of the knowledge bases when determining and linking relations from text. Therefore, we further extend this model by introducing knowledge integration and validation mechanisms. Knowledge integration enhances the encoder representation by infusing structured data from the KB, consisting of a set of relation candidates connected with the entities pre-identified in the questions. Such knowledge integration can have a two-fold advantage: (a) enhancing the performance of the relation linking model when there is a lack of training data by using information from the knowledge graph; (b) ability to deal with unseen relations since it is transformed into a re-ranking task.

The main contributions of this work are as follows: 
\begin{itemize}
    \item a novel generative model for relation linking in the context of KBQA;
    \item a knowledge integration that enhances the model with information from the knowledge base to handle unseen relations and a  knowledge validation module to further filter, disambiguate and re-rank the relations generated by the seq2seq model;
    \item an extensive experimental evaluation on four KBQA datasets showing large improvements over the state-of-the-art. We obtain an F1 increase between $9\%-59\%$  over the state-of-the-art on different datasets derived from knowledge bases such as DBpedia and Wikidata.
\end{itemize}

\section{GenRL: Generative Relation Linking}
\label{sec:model}

\begin{figure}[t]
  \centering
    \includegraphics[width=0.9\textwidth]{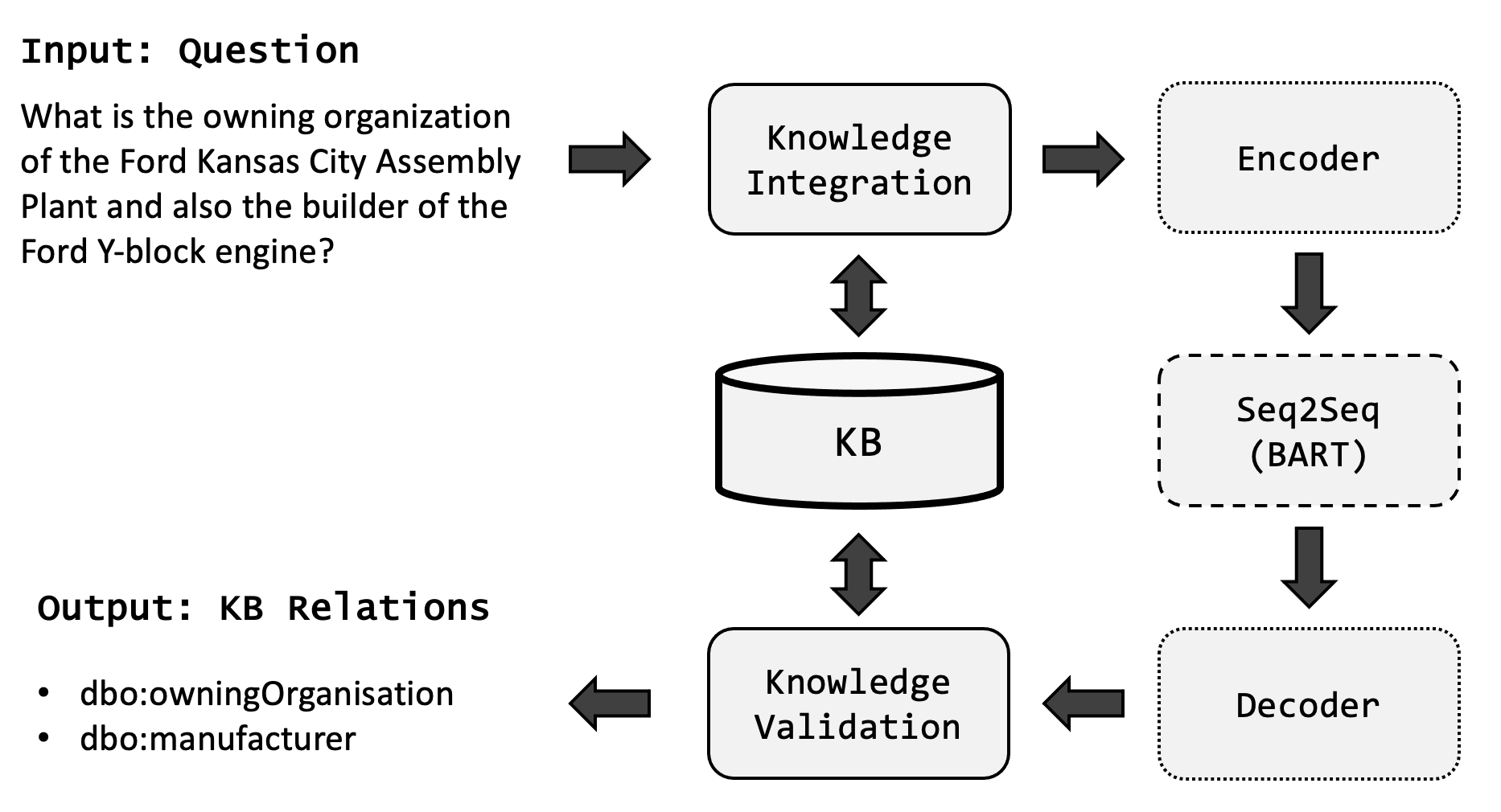}
    \caption{GenRL Framework}
    \label{fig:genrl}
\end{figure}

In this section, we describe GenRL, our generative method for relation linking. Our approach is based on an encoder-decoder paradigm where a model is trained to transform a sequence of input tokens into a sequence of target tokens. 
Formally, let us define $S=[s_1,...,s_N]$ as the source sequence given as input to the encoder, and $T=[t_1,...,t_M]$ as the target sequence generated by the decoder.

The probability of the target sequence is defined as: $P(T|S) = \prod_{k=1}^M(P(t_k|t_{<k},S))$.

The probability of generating the token $t_k$ at step $k$ is conditioned on the entire source sequence as well as the tokens that have been generated so far by the decoder on the target side. In a straight forward application of seq2seq models for relation liking, the input would be the  question text and the output would be a sequence of KB relations. In GenRL, we adopt BART \cite{DBLP:conf/acl/LewisLGGMLSZ20}, a pre-trained seq2seq language model based on the transformer \cite{DBLP:conf/nips/VaswaniSPUJGKP17} architecture, with two main components: a bi-directional encoder and a left-to-right decoder. 
BART achieves remarkable performance when fine-tuned on sequence generation tasks, making it a good candidate for our problem. 

Figure \ref{fig:genrl} shows a high-level overview of the GenRL architecture. The system takes a natural language question as input.
A necessary first step in our approach is to recognise the entities in the question and link them to the target KB using an entity linking system. 
The Knowledge Integration module (Section \ref{sec:encoder}) aims to query the KB, enrich the question with a list of candidate relations according to the detected entities, and prepare the encoder representation for the seq2seq model. 
The decoder of seq2seq model generates a structured sequence consisting of a list of argument-relation pairs, based on the enriched input representation (see Section \ref{sec:decoder} for details). 
Finally, the Knowledge Validation module (Section \ref{sec:kv}), analyses the top-k most probable relation sequences generated by the model, and uses the argument values for the relations in the sequence to determine if the sequence is consistent with the KB content.

\subsection{Encoder Input Representation}
\label{sec:encoder}
\begin{figure}[t]
  \centering
    \includegraphics[width=0.95\textwidth]{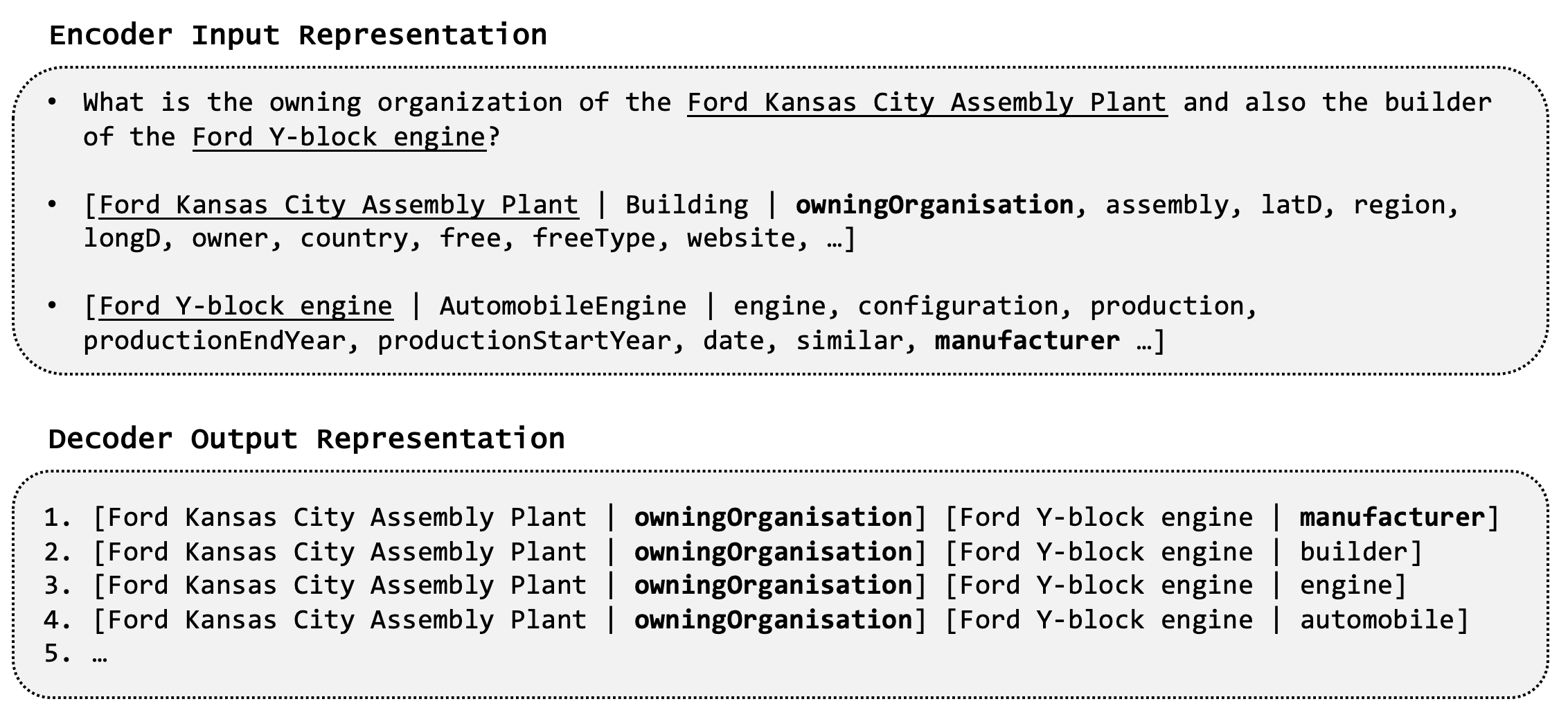}
    \caption{Input-Output representations for the sequence-to-sequence model}
\label{fig:enc_dec}
\end{figure}

Given a question as input, the Knowledge Integration module extracts additional information from the KB to prepare the encoder representation for the seq2seq model, as shown in Figure \ref{fig:enc_dec}.
In order to allow access to the KB, we first identify and link the entities in the question using an entity linker. In our case, we used BLINK combined with a neural mention detection model~\cite{wu2019zero}. 
For each linked entity, we build a text structure comprised of the entity mention in question, the entity type defined in the KB ontology and a list of relations\footnote{In the encoder-decoder representations, we consider only the relation names or labels by removing the URIs and namespaces for DBpedia and converting the property ID to the corresponding relation label for Wikidata. The knowledge validation module converts the relation labels back to URIs.} directly connected with the entity: \texttt{[Entity mention | Entity type | Rel1, ..., RelN]}. The entity structure for all entities in the question is concatenated with the natural language question. When an entity is typed with multiple classes, we use the class hierarchy information to find the most specific type that will prune all the generic types. If there are more than one classes after pruning, the class with most instances in the KG is used. 

This new representation has three advantages: 1) it provides detected entities explicitly to the model; 2) enriches the encoder with local information about the entities in the question, such as their types; 3) it provides a pre-built list of relations used as possible candidates.
With this enriched representation, we observe an increased generalisation capability of the seq2seq model showing better performance. Moreover, this representation assists the model in generating relations that have not been seen during training by exposing the model to a list of candidate relations from the KB. This is helpful especially for those relation type labels which have a lexical gap with the text in the question.

However, BART's encoder can handle only a limited number of tokens (i.e 512) and the entity data structures may  exceed this limit when there is a high number of distinct relations connected to the entities. In order to address this issue, we pre-rank the relations for each entity in the question using the word embedding similarity technique between the question and the relation labels similar to the lexical similarity approach described in \cite{mihindukulasooriya2020leveraging}.

\subsection{Decoder Output Representation}
\label{sec:decoder}
We design the target sequence for the decoder using a data structure formatted as follows: \texttt{[Arg1 | Rel1], ..., [ArgN | RelN]}. For each predicted relation, the model also generates one of its arguments. 
The relation arguments can be KB entities that appear in the question, or placeholders for answer variable or unbound intermediate variables for multi-hop relations in the query.
In the first case, we train the seq2seq model in order to generate the entities recognised in the question paired with the corresponding KB relations.
In the example in  Figure~\ref{fig:enc_dec}, the model generated entity \textit{Ford Kansas City Assembly Plant } as an argument for the relation \textit{owningOrganisation} and the entity \textit{Ford Y-block engine} as an argument for the relation \textit{manufacturer}.

In the second case, the model generates placeholders for unbound variables. We show such an example in see Figure \ref{fig:kbval2}, where the relation \textit{dbo:owner} is not directly connected with any entities in the question.
Our strategy is to pair these multi-hop relations with the question Wh terms (i.e. \textit{Who}) used as a placeholder. We use the gold SPARQL queries in the training set to generate this output for training the model.

\subsection{Knowledge Validation}
\label{sec:kv}
During  knowledge validation the system analyses each candidate output sequence produced by the decoder.  In this phase we map the arguments (entity mentions or Wh terms) back to entity URIs or variables, use them to validate candidate outputs and convert the relation labels into URIs in the KB ontology with the correct namespaces. 

We collect all the argument-relation pairs for a given output sequence and build all possible graphs that are subsequently used to query the KB. If one of the  resulting graphs is matched in the KB then we consider the predicted sequence is valid. We discard the sequences that the model produces in cases when none of the graphs is matched in the KB.
Building all possible graphs based on the argument-relation pair uses the following set of heuristics: 

\begin{figure}[t]
  \centering
    \includegraphics[width=1\textwidth]{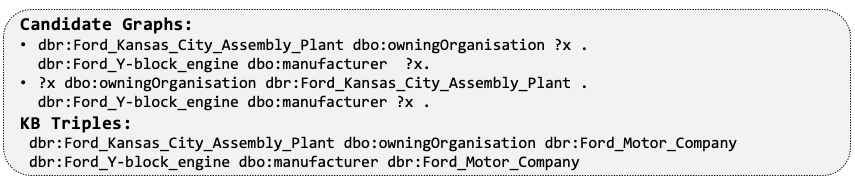}
 \caption{Knowledge Validation example for a sequence of entity-relation pairs. This shows how the first decoder output sequence from Figure~\ref{fig:enc_dec} is validated.}
\label{fig:kbval1}
 \end{figure}

\subsubsection{Entity-Relation Heuristics.} We expand each entity-relation pair into triples by first considering the possible namespaces for the predicted relation labels. For the case of DBpedia, the namespaces are \textit{dbo:}\footnote{\url{http://dbpedia.org/ontology/}} and \textit{dbp:}\footnote{\url{http://dbpedia.org/property/}}. Next, we consider two triples where the entity is either in the subject or object position. To complete the triple, we use an unbound variable \textit{?x} to indicate the missing argument. To create a single connected graph, entity-relation pairs in the same candidate sequence use the same unbound variable \textit{?x} across all triples. Each entity-relation pair creates four triples and cartesian product of triples from each entity-relation pair creates all possible candidate graphs. In order to make this process efficient, we prune the invalid single triples first before expanding with product to create candidate graphs. Furthermore, it follows decoder ranking and stops as soon as the first valid candidate graph is found. Finally in Figure~\ref{fig:kbval1} we show two possible candidate graphs for a given model output. The first graph has a match in the KB which validates the sequence produced by the model. The KB triples that match the first graph are shown at the bottom of Figure~\ref{fig:kbval1}. 
In the example in Figure~\ref{fig:enc_dec}, we validate the decoder sequence on the first position. In cases where the first generated relation sequence can not be validated against the KB (because none of its graphs is matched), we proceed to the next generated sequence and the process stops as soon as the first valid sequence is found.

\subsubsection{Placeholder-Relation Heuristics.} 

We expand each placeholder-relation pair into triples similarly to entity-relation pairs. In this case, the placeholder is replaced with a new unbound variable \textit{?y} to represent the \textit{unknown} or the \textit{answer}. We complete the triple with the unbound variable \textit{?x} similar to the previous case to connect the triples to each other and create a set of candidate graphs.
 
\begin{figure}[t]
  \centering
    \includegraphics[width=0.95\textwidth]{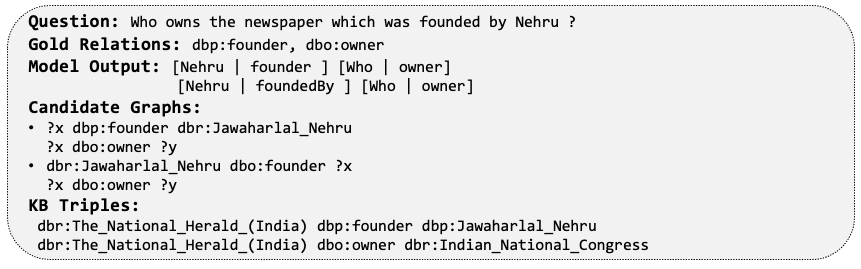}
 \caption{Knowledge validation example for a sequence of entity-relation and placeholder-relation pairs}
\label{fig:kbval2}
 \end{figure}

In the example in Figure~\ref{fig:kbval2} we show two possible graphs for the first sequence generated by the model, with one placeholder-relation pair. The first graph is matched by the KB content and we show the matching triples in the figure. Since at least one of the graphs we produced for the output sequence has been matched, the output system is valid and relation labels can be converted to their corresponding URIs. It is worth noting though that this process of only using two unbound variables does not scale well to arbitrarily long questions with a large number of triples and we plan to investigate it as our future work. 

We validate the top $N$ query candidates according to the ranking order of the decoder sequentially ($N$ = 50 in our experiments). The KG validation phase stops once we find a valid candidate query graph with matching triples in the KB. Thus, if there are other valid graphs with lower confidence at lower ranks in the decoder, they will be automatically ignored.

In the previous example we explained the process using DBpedia as the KB. As for Wikidata, we have followed a similar process but due to complexities of the Wikidata model, it requires handing reified statements and qualifier properties using several other patterns. In contrast to DBpedia, relations can be either connected to entities directly (\textit{wdt:}\footnote{\url{http://www.wikidata.org/prop/direct/}}) or through reified statements (\textit{p:}\footnote{\url{http://www.wikidata.org/prop/}},\textit{ps:}\footnote{\url{http://www.wikidata.org/prop/statement/}}, \textit{pq:}\footnote{\url{http://www.wikidata.org/prop/qualifier/}}). For example, qualifier relations are only associated with statements and some specific relations such as ``\textit{instance of (P31)}" or ``\textit{subclass of (P279)}" is only attached to entities and not statements. Once all SPARQL query variations are generated according to the Wikidata model, the validation process is similar to one described for DBpedia.   

KBQA datasets contain \textit{ASK} questions that have to be treated differently because, by design, when the expected answer is false such as ``Was Barack Obama president of Canada?", these questions contain triple patterns that are not present in the KG. 
We handle this by using two simple heuristics (a) Identify ASK questions using the question tokens, and (b) train the decoder argument pairs for \textit{ASK} to be ``[E1 - RelA] [E2 - RelA]". 

Once an \textit{ASK} query is detected, GenRL relaxes the KV to adapt to possible \textit{false ASK} questions using the following strategy. In particular, we first try to validate top $N$ decoder outputs ($N$=10, in our experiments) assuming it's an ASK question with a True answer (\textit{i.e.}, a valid triple in the KB). Generally, as ASK triples have both entities bounded, a positive validation gives a stronger signal. If none of the top n candidates are validated with KG, we return the top decoder output assuming it's an ASK question with a NO (False) answer.

\section{Evaluation}
\label{sec:evaluation}

In this section, we detail our experimental setup and evaluate our approach against the state-of-the-art KBQA relation linking approaches. We adopt standard evaluation metrics such as precision, recall, and F1 on DBpedia and Wikidata based KBQA datasets. 

\subsection{Experimental Setup}
\subsubsection{Benchmarks} We perform experiments on four datasets targeting two popular KBs, DBpedia and Wikidata. Each question in these datasets comes with its corresponding SPARQL query, annotated with gold relations. In particular, we used the following datasets:
\begin{itemize}
    \item \textbf{QALD-9}~\cite{DBLP:conf/semweb/UsbeckGN018}: is a dataset based on the DBpedia (2016-04 version) with 408 training questions and 150 test questions in natural language. The questions and the gold SPARQL queries are manually created.
    \item \textbf{LC-QuAD 1.0}~\cite{lcquad}: is another dataset based on DBpedia (2016-10 version) with a total of 5,000 questions (4,000 train and 1,000 test) based on templates and then paraphrased. 
    \item \textbf{LC-QuAD 2.0}~\cite{lcquad2}: A large dataset based on Wikidata with 6,046 test questions and around 24k training questions. Questions in this dataset have a good variety and complexity levels such as multi-fact questions, temporal questions and questions that utilise qualifier information.
    \item \textbf{SimpleQuestions-WD}~\cite{wikidata-benchmark}: A version of the popular SimpleQuestions dataset mapped to Wikidata. It comprises of 5,622 test questions, and around 19K training questions. This is a subset of the original dataset on Freebase which contained 108K questions. As the name implies, all questions in this dataset are simple with queries encompassing a single triple in the KB. 
\end{itemize}

\subsubsection{Baselines}
For the DBpedia-based benchmarks, we compare {\sysname} with Falcon~\cite{sakor2019old} and SLING \cite{mihindukulasooriya2020leveraging}. As for Wikidata-based benchmarks,  we compare against Falcon 2.0 \cite{sakor2020falcon} and KB-Pearl \cite{lin2020kbpearl}. We did not directly compare with the other systems on SimpleQuestions (Freebase) such as Lukovnikov et al.~\cite{DBLP:conf/semweb/LukovnikovF019} (F1: 0.83) because SimpleQuestions(Wikidata) is on a different KG and is a smaller subset. Finally, we provide a seq2seq baseline (\sysname~wo/KB)  by fine-tuning BART having only the question as a source and the list of relations as a target.

\subsubsection{Model settings}
We trained our seq2seq model using BART-large on the training data provided for each dataset and set the encoder size to 512 tokens. We used 2 NVIDIA V100 GPUs to train the models over 10 epochs with a batch size of 4. With this setup, the models generally do not require long training time. For example, on LC-QuAD 2.0, the largest dataset, the training requires 12hrs. On QALD-9, with a few hundred examples, the train runtime is only 9 minutes.
During inference, we expanded the beam search up to 50 beams in order to generate the top-50 list of entity-relation pairs ranked by their probabilities.

\subsection{Results}
Table~\ref{tab:dbpedia_results} and Table~\ref{tab:wikidata_results} show the results of {\sysname} in comparison to other state-of-the-art approaches on DBpedia and Wikidata based datasets. These results evidently show that {\sysname} outperforms all the existing approaches by a large margin, i.e. achieving a higher F1 score between 9 points (compared to SLING on QALD) and 59 points (compared to Falcon on Simple Questions-WD). 

The results, particularly for BART, show that vanilla seq2seq models in most cases perform better than the state-of-the-art relation linking approaches such as SLING, Falcon, and KBPearl. This clearly demonstrates that the challenges with relation linking can be naturally addressed using simple seq2seq models. Furthermore, our model {\sysname} is using knowledge integration and performs better than the baseline seq2seq model on all the datasets. These results show the positive impact of the KB integration in {\sysname}, which we further demonstrate with extensive analysis and ablation study in the next sections.

\begin{table}[t]
\centering
\bgroup
\setlength{\tabcolsep}{5pt}
\begin{tabular}{l| c c c | c c c} 
\toprule
 & \multicolumn{3}{|c}{\textbf{LC-QuAD 1.0}} & \multicolumn{3}{|c}{\textbf{QALD-9}} \\ 
 \hline
& \textbf{P} & \textbf{R} & \textbf{F1} & \textbf{P} & \textbf{R} & \textbf{F1} \\ 
\midrule
\textbf{Falcon 1.0}~\cite{sakor2019old} & 0.42 & 0.44 & 0.43\tablefootnote{These numbers differ from the cited paper because we only performed evaluation on the test set in this experiment setup. The cited papers used both training and test set for their evaluation. We reevaluated them only for test set.\label{note:test_set}} & 0.23 & 0.23	& 0.23\textsuperscript{\getrefnumber{note:test_set}}\\
\textbf{SLING}~\cite{mihindukulasooriya2020leveraging} & 0.41 & 0.55 & 0.47\textsuperscript{\getrefnumber{note:test_set}}  & 0.39 & 0.50 & 0.44\textsuperscript{\getrefnumber{note:test_set}} \\ 
\midrule
\textbf{GenRL wo/KB} & 0.47 & 0.50 & 0.48 & \textbf{0.51} & 0.43 & 0.47 \\
\textbf{GenRL} & \textbf{0.54} & \textbf{0.74} & \textbf{0.60} & 0.49 & \textbf{0.61} & \textbf{0.53} \\ 
\bottomrule
\end{tabular}

\egroup
\caption{Relation linking results on DBpedia based datasets. GenRL wo/KB refers to our model without Knowledge Integration and Knowledge Validation.}
\label{tab:dbpedia_results}
\end{table}

\begin{table}[t]
\centering
\bgroup
\setlength{\tabcolsep}{5pt}
\begin{tabular}{l| ccc | ccc | ccc} 
\toprule
 & \multicolumn{3}{|c|}{\textbf{LC-QuAD 2.0}} & \multicolumn{3}{|c|}{\textbf{LC-QuAD 2.0$_{1942}$}} & \multicolumn{3}{|c}{\textbf{SimpleQ WD}} \\ \hline
& \textbf{P} & \textbf{R} & \textbf{F1} & \textbf{P} & \textbf{R} & \textbf{F1} &  \textbf{P} & \textbf{R} & \textbf{F1} \\\hline

\textbf{Falcon 2.0}~\cite{sakor2020falcon} & 0.44 & 0.37 & 0.40 & 0.43 & 0.32 &   0.36\tablefootnote{We calculated the results for the subset using the file at \url{https://github.com/SDM-TIB/falcon2.0/blob/master/datasets/results/test\_api/falcon\_lcquad2.csv}} & 0.35 & 0.44 & 0.39 \\ 
\textbf{KBPearl}~\cite{lin2020kbpearl} & - & - & - & 0.57 & 0.48 & 0.52\tablefootnote{The KBPearl paper reports F1 of 0.41 due to a typo but its authors confirmed the correct F1 to be 0.52.}  & - & - & - \\
\midrule
\textbf{GenRL wo/KB} & 0.81 & 0.81 & 0.81 & 0.87 & \textbf{0.86} & 0.87 & 0.96 & 0.96 & 0.96 \\
\textbf{GenRL} & \textbf{0.88} & \textbf{0.82} & \textbf{0.84} & \textbf{0.89} & 0.85 & \textbf{0.87} & \textbf{0.98} & \textbf{0.98} & \textbf{0.98} \\ 
\bottomrule
\end{tabular}
\egroup
\caption{Relation linking results on Wikidata based datasets. LC-QuAD 2.0$_{1942}$ is the subset used by KBPearl~\cite{lin2020kbpearl}.}
\label{tab:wikidata_results}
\end{table}

\subsection{Detailed Analysis}

\subsubsection{Accuracy of predicting the number of relations} In order to evaluate the system's ability to predict the correct number of relations, we have calculated the percentages of questions where (a) predicted number of relations is same as the number of gold relations, (b) predicted number of relations is larger than the gold relations and (c) the predicted number of relations is smaller than the number of gold relations. This experiment checks only the accuracy of predicting the correct number of relations, without considering if relations themselves are correct. Table~\ref{tab:number_of_triples} indicates that seq2seq models are stronger in predicting the correct number of relations from text compared to rule-based systems such as Falcon 1.0 and 2.0. GenRL wo/KB model has slightly better performance in predicting the correct number of relations. In our analysis, the slight decrease was mainly influenced by the entity linking error propagation during KI. Furthermore, we can see that all systems perform better on template-based datasets (LC-QuAD 1.0 / 2.0) than manually constructed datasets (QALD-9).

\begin{table}[t]
\centering
\begin{tabular}{l|ccc|ccc|ccc}
\hline
\multicolumn{1}{c|}{\textbf{Dataset}} & \multicolumn{3}{c|}{\centering \textbf{QALD - 9}} & \multicolumn{3}{c|}{\centering \textbf{LC-QuAD 1.0}} & \multicolumn{3}{c}{\centering \textbf{LC-QuAD 2.0}} \\ \hline
\multicolumn{1}{c|}{Num of rels} &
  \textbf{\begin{tabular}[p{1cm}]{@{}c@{}}pred  \\ = \\ gold\end{tabular}} &
  \textbf{\begin{tabular}[c]{@{}c@{}}pred \\ \textgreater\\  gold \end{tabular}} &
  \textbf{\begin{tabular}[c]{@{}c@{}}pred  \\ \textless\\ gold\end{tabular}} &
  \textbf{\begin{tabular}[c]{@{}c@{}}pred \\  = \\ gold \end{tabular}} &
  \textbf{\begin{tabular}[c]{@{}c@{}}pred \\          \textgreater\\ gold \end{tabular}} &
  \textbf{\begin{tabular}[c]{@{}c@{}}pred  \\           \textless\\ gold \end{tabular}} &
  \textbf{\begin{tabular}[c]{@{}c@{}}pred  \\ = \\  gold \end{tabular}} &
  \textbf{\begin{tabular}[c]{@{}c@{}}pred \\ \textgreater\\ gold\end{tabular}} &
  \textbf{\begin{tabular}[c]{@{}c@{}}pred \\ \textless\\ gold \end{tabular}} \\ 
\midrule
Falcon 1/2                           & 26\%       & 23\%       & 51\%      & 43\%        & 34\%        & 23\%       & 31\%          & 16\%        & 54\%        \\ 
GenRL wo/KB                                   & 70\%       & 9\%        & 21\%      & 93\%        & 1\%         & 6\%        & 94\%         & 1\%        & 5\%        \\ 
{\sysname}                                  & 69\%        & 7\%        & 24\%      & 87\%        & 1\%         & 12\%       & 92\%         & 1\%        & 7\%        \\ 
\bottomrule
\end{tabular}
\caption{A comparison of the predicted number of relations vs the number of gold relations in the LC-QuAD 2.0 dataset.}
\label{tab:number_of_triples}
\end{table}

\subsubsection{Entity linking error propagation} 
In order to understand the impact of entity linking which is used by both knowledge integration and validation steps, we performed an experiment on LC-QuAD 1.0 using gold standard entities similar to EERL~\cite{pan2019entity}. EERL reported an F1 of 0.55 with a precision of 0.53 and a recall of 0.58. With gold entities, {\sysname} resulted in an F1 of 0.68 with a precision of 0.60 and a recall of 0.83 compared to the 0.60 F1 with machine entity linking. Gold entities help to align the questions better with KG in both Knowledge Integration (improving recall) and Knowledge Validation (improving precision).

\subsubsection{Impact on end-to-end KBQA performance} In order to check the impact on KBQA, we have used the state-of-the-art KBQA system by \cite{kapanipathi2020question} and replaced its relation linking module with {\sysname}. For LC-QuAD 1.0, it results in a $\sim$15\% point increase in Macro F1 from 44.45 to 59.63. We intend to investigate this further and expand it to other datasets in the future.

\subsection{Error Analysis}
\label{sec:error_analysis}

\paragraph{\textbf{LC-QuAD 1.0}}

While analysing the low precision of our results in LC-QuAD 1.0 dataset, we noticed that the dataset used for this benchmark, that is, DBpedia 2014-04 version has an issue of redundancy in relations. For example, \textit{Ben Ysursa} and \textit{Gonzaga University} are connected using both \textit{dbo:almaMater} and \textit{dbp:almaMater} relations. In such cases, the gold standard query can contain either one of them. It is not possible for relation linking systems to produce the exact relation in terms of \textit{dbo:}/\textit{dbp:} variant as in the gold standard since both of them are equally valid (in terms of retrieving the same exact answer from the KB). For example, Table~\ref{tab:lcquad_false_negatives} shows a question with its gold relation set compared to three other equally valid relation sets where each one of them gets a different F1 score according to how much it matches the specific set of gold relations.  

\begin{table}[t]
\centering
\begin{tabular}{l|c|c}
\hline
\textbf{Gold Standard Query} & \begin{tabular}[c]{@{}l@{}}\textbf{Relations yielding}\\ \textbf{the same answer}\end{tabular} &  \begin{tabular}[c]{@{}l@{}} \textbf{Rel Prediction} \\ \textbf{F1} \end{tabular}\\ \hline
\multirow{4}{*}{\begin{tabular}[c]{@{}l@{}}In which state is the alma mater of Ben Ysursa \\ located?\\ \\  SELECT DISTINCT ?uri WHERE \{ \\   dbr:Ben\_Ysursa dbp:almaMater ?x . \\  ?x dbo:state ?uri  . \\ \}\end{tabular}} &
  \begin{tabular}[c]{@{}l@{}}dbp:almaMater\\ dbo:state\end{tabular} &
  1.0 \\ \cline{2-3} 
                    & \begin{tabular}[c]{@{}l@{}}dbo:almaMater\\ dbo:state\end{tabular}     & 0.5               \\ \cline{2-3} 
                    & \begin{tabular}[c]{@{}l@{}}dbp:almaMater\\ dbp:state\end{tabular}     & 0.5               \\ \cline{2-3} 
                    & \begin{tabular}[c]{@{}l@{}}dbo:almaMater\\ dbp:state\end{tabular}     & 0.0               \\ \hline
\end{tabular}
\caption{An example query from LC-QuAD 1.0 training set}
\label{tab:lcquad_false_negatives}
\end{table}

In order to understand the significance of the problem, we have analysed the 4,000 training questions in LC-QuAD 1.0 and found 2,623 (66\%) of them had other variations of valid queries (queries that will generate non-empty results) only by changing the namespace (e.g., \textit{dbo:state} vs \textit{dbp:state}). In 1,587 variations, they produced the exact same list of answers as the query in the gold standard and in 881 cases they produced a partial match with the gold answer, and in 155 cases they produced a different answer. If we create all valid SPARQL query variations based on the answer set overlap and re-evaluated our system allowing any of those equivalent combination to be the gold query, GenRL gets an F1 of 0.73 (P: 0.72 and R: 0.76) compared to the standard evaluation of 0.60 F1. This provides evidence that the precision of GenRL on LC-QuAD 1.0 in Table~\ref{tab:dbpedia_results} is affected by this issue of the DBpedia KB.

\begin{figure}[t]
    \centering
    \includegraphics[width=0.95\textwidth]{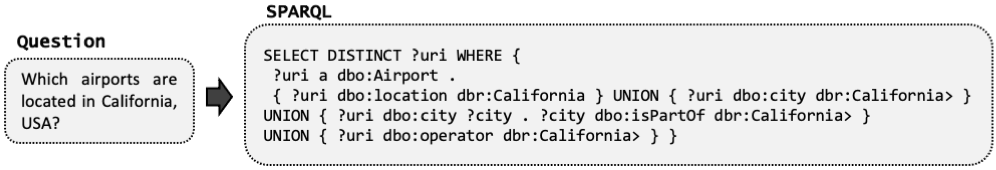}
    \caption{A SPARQL query using UNIONs from QALD-9 dataset. Here the single relation \textit{located in} is mapped to four KB relations: \textit{location}, \textit{city}, \textit{isPartOf} and \textit{operator}.}
    \label{fig:complex_queries}
\end{figure}

\paragraph{\textbf{QALD-9}} This dataset contains complex queries that sometimes contain several unions to fit exactly to the question that is being asked and the KB content as shown in Fig~\ref{fig:complex_queries}. Predicting relations for such complex queries is challenging for all relation linking systems. 

\paragraph{\textbf{LC-QuAD 2.0}} We noticed that gold SPARQL queries contained some relations that are deleted\footnote{\url{https://www.wikidata.org/wiki/Wikidata:Requests\_for\_deletions/Archive/2019/Properties/1}} such as \textit{P134}, \textit{P727}, and \textit{P1112}. In LC-QuAD 2.0 training data, we counted 20 such relations. Our evaluation was run on a snapshot of April, 2021 version of the KB and Wikidata has evolved significantly since 2019, the time LC-QuAD 2.0 was created. Nevertheless, we assume that most facts in questions might not have changed and the negative impact of this on reported numbers to be minimal. Furthermore, we have noticed that some of the questions do not match with their SPARQL queries. For example, there were some questions with text such as ``What is it?" or ``How is it". 

Finally, we observed some unnatural questions due to the use of templates, e.g., ``Who is the country for head of state of Mahmoud Abbas?". Despite these issues, {\sysname} was able to outperform all existing systems and achieves a promising performance across all datasets. This indicates that {\sysname} is tolerant to these types of questions. 

\section{Discussion}
\label{sec:discussion}

\subsection{Qualitative Analysis}
In this experiment, we took a random 10\% of LC-QuAD 1.0 training data as a training subset and another 10\% as a validation set with the number of unseen relations in the validation set being 114 relations. Table~\ref{qualitative_analysis_tab} shows a number of examples from the validation dataset. {\sysname} could predict relations where there is a lexical gap between the question text and the relations itself such as \textit{settlementType} and \textit{placeOfBurial}. It was also able to predict multiple explicit (e.g. \textit{network}, \textit{sire}),  implicit (e.g. \textit{honours}, \textit{starring}), and even unseen relations (e.g. instrument and cpu) thanks to its knowledge integration and validation steps. However, implicit relation and relations with lexical gap still pose a challenge on {\sysname} and on all existing relation linking approaches. In particular, for the question \textit{Name the rivers who originate from Essex?}, the question text does not imply why a model would prefer ``mouthPlace" (gold) over \textit{sourceRegion} (predicted). Similarly, in the question \textit{Who acted in the movies whose music is composed by Walter Scharf?}, again the text for ``acted" is actually closer to the predicted relation ``starring" than to the gold ``artist". We intend to investigate further on how to use KB knowledge to handle such cases in our future work.

\begin{table}[t]
 \resizebox{1\columnwidth}{!}{
\begin{tabular}{l|c|c|c}
\toprule
\textbf{Question} & \textbf{Gold} & \textbf{Predicted} & \textbf{Correct} \\
\midrule
\textbf{Single Relation}: & & & \\
What are the towns who have Thesaban system? & settlementType & settlementType & \cmark\\
Where is the grave of Ivan III of Russia? & placeOfBurial & placeOfBurial & \cmark\\
\midrule
\textbf{Multiple Relations}: & & & \\
In which sitcom did Jeff Conaway acted &&&\\and had TNT as its network? & starring, network & starring, network & \cmark \\
Which awards have been given to the horse &&&\\ who sired Triplicate?& sire, honours& sire, honours & \cmark \\
\midrule
\textbf{Unseen Relations}: & & & \\
 What famous musicians play the remo?  & instrument & instrument & \cmark \\
 Which appliance's CPU is Cell (microprocessor)&&&\\and predecessor is PlayStation 2? & cpu, predecessor & cpu, predecessor & \cmark \\
\midrule
\textbf{Wrong Predictions}: & & & \\
Name the rivers who originate from Essex?& mouthPlace & sourceRegion & \xmark \\
Who acted in the movies whose music is &&&\\composed by Walter Scharf? & musicComposer, artist & musicComposer,  starring & \xmark \\
\bottomrule
\end{tabular}
}
\caption{Qualitative Analysis of {\sysname} predictions from LC-QuAD-1 dataset}
\label{qualitative_analysis_tab}
\end{table}

\subsection{Generative Structured Output Evaluation}
\label{sec:seq2seq_eval}

\begin{table}[t]
\centering
\bgroup
\setlength{\tabcolsep}{5pt}
\def\arraystretch{1.2}
\begin{tabular}{l|c|ccc} 
\toprule
& n. train  & \textbf{P} & \textbf{R} & \textbf{F1}  \\ \hline
\textbf{QALD-9} &  398 & 0.65 & 0.63 & 0.63   \\ 
\textbf{LC-QuAD 1.0} &  4,000 & 0.73 & 0.76 & 0.74   \\
\textbf{LC-QuAD 2.0} & 24,000 & 0.85 & 0.86 & 0.85  \\
\textbf{SimpleQ WD} & 19,235  & 0.98 & 0.98 & 0.98  \\
\bottomrule
\end{tabular}
\egroup
\caption{Results for the structured output generated by the seq2seq model}
\label{tab:seq2seq_results}
\end{table}

Table \ref{tab:seq2seq_results} shows the results computed only considering the output from the seq2seq model using the argument-relation representation as the gold standard. 
On DBpedia-based datasets, we observe higher numbers compared to the results of {\sysname} showed in Table \ref{tab:dbpedia_results} (+10 F1 on QALD-9, +14 F1 on LC-QuAD 1.0). In this case the seq2seq model has been trained on relation labels without URIs. The difference in performance can be explained  by the challenge of disambiguating the appropriate namespaces (\textit{dbo} vs \textit{dbp}) as discussed in Section~\ref{sec:error_analysis}.
It is worth noticing the performance achieved on QALD-9 despite the fact that the model has been fine-tuned only on 398 examples.
On both Wikidata-based datasets, we observe very high numbers mainly due to the availability of larger training sets. In particular, the seq2seq model pushes the boundaries on SimpleQuestions-WD obtaining an F1 of around 98\% solving the task for this dataset.

\subsection{Training with Less Data}
\label{sec:learning_curve}
In this section, we study the performance of the system on LC-QuAD 1.0, as we vary the size of the training set. We hold out a subset of randomly selected 400 questions from the training set that we use as a development set. We create different training splitting on the remaining part.

\begin{table}[t]
\centering
\bgroup
\setlength{\tabcolsep}{5pt}
\def\arraystretch{1.2}

\begin{tabular}{l|c c c} 
\toprule
\textbf{Train (\%)} & \textbf{P} & \textbf{R} & \textbf{F1} \\ \hline
1\%   & 0.53 & 0.47 & 0.48  \\
10\%  & 0.64 & 0.66 & 0.63  \\
20\%  & 0.68 & 0.72 & 0.69  \\
40\%  & 0.73 & 0.78 & 0.74  \\
60\%  & 0.75 & 0.80 & 0.77  \\
80\%  & 0.77 & 0.82 & 0.78  \\

\bottomrule
\end{tabular}

\egroup
\caption{Training with less data study on LC-QuAD 1.0, {\sysname} trained on a percentage of training data and tested on a development set of 400 questions}
\label{tab:learning_curve}
\end{table}

Table \ref{tab:learning_curve} reports the results of this study. Each row shows the performance of {\sysname} trained on different portions of the original training set. Surprisingly, the model trained only on 1\% of the training set (i.e. 40 examples) obtains 48\% F1. In addition, with the 20\% the model achieves performance close to that obtained by a fully trained model.

\section{Related Work}
\label{sec:related}

Knowledge base question answering has become a popular task due to its relevance to many real-world applications. The recent KBQA systems, particularly on knowledge bases such as DBpedia~\cite{auer2007dbpedia} and Wikidata~\cite{vrandevcic2014wikidata} can be categorized into rule-based, unsupervised systems~\cite{kapanipathi2020question,hu2017answering} and end-to-end trained models~\cite{maheshwari2019learning,chen2020formal,DBLP:conf/acl/YuYHSXZ17}. Rule-based approaches~\cite{kapanipathi2020question,hu2017answering} use semantic/dependency parses and have shown to be highly effective for KBQA. Among supervised approaches pre-trained language models have been popularly used for answering questions over a knowledge base. 
In both of these categories of KBQA systems, the performance of transforming natural language question text to SPARQL is impacted by entity and relation linking components~\cite{kapanipathi2020question}. In particular, relation linking has shown to be the primary error propagation module and needs to be significantly improved. 

 Existing relation linking approaches can be broadly categorised into rule-based, distantly supervised and strictly supervised methods. Several rule-based systems have been proposed recently for relation linking ~\cite{sakor2019old,pan2019entity,earl,rematch,sakor2020falcon}. Among those, Falcon~\cite{sakor2019old} jointly links entities and relations in a question to DBpedia using a sequence of steps including POS tagging, n-gram tiling and compounding. Falcon 2.0~\cite{sakor2020falcon} is the recent version of Falcon that performs linking to Wikidata knowledge base. Similarly, Entity Enabled Relation Linking (EERL)~\cite{pan2019entity} investigated the use of questions' entities to support relation linking task over DBpedia KB. KBPearl~\cite{lin2020kbpearl} is another system that performs joint entity and relation linking to Wikidata. It first creates a semantic graph of text using OpenIE and maps both entities and relations to a given KB.
SLING~\cite{mihindukulasooriya2020leveraging} is an example of a distantly supervised system. It leverages semantic parsing techniques for better question understanding and builds an ensemble of approaches (e.g., statistical mapping, word embedding) to achieve state-of-the-art performance on various DBPedia datasets. Among those components, a BERT-based distantly supervised relation extraction system is trained using sentences automatically collected from Wikipedia. Compared to these approaches, GenRL has the important advantage of not being KB-specific, which enables easy domain portability across different KBs. In addition, GenRL does not require the use of NLP components such as semantic parsing that helps reduce error propagation in the overall approach.

\section{Conclusions and Future Work}
\label{sec:conclusions}
In this work, we show that relation linking can be formulated as a sequence generation problem leveraging recent advancements in auto-regressive sequence-to-sequence models. This simple yet powerful approach is shown to largely outperform all existing relation linking systems that apply sophisticated heuristics over several datasets. 
To further improve this model, we proposed the knowledge integration and validation strategies which infuse the structure of the underlying  knowledge base into the neural model. In our experiments, we show that this strategy helps the model to better generalise especially on relations not previously seen during training. The knowledge integration and validation steps resulted in absolute improvements of up to 12\% on F1 score compared to the simple seq2seq model. In our research agenda, we plan to investigate generative models with knowledge integration to model the end-to-end KBQA setup.

\bibliographystyle{unsrtnat}
\bibliography{references}

\end{document}